\documentclass[submission,copyright,creativecommons]{eptcs}
\providecommand{\event}{ICLP \& LPNMR DC 2024} 

\usepackage{iftex}

\ifpdf
  \usepackage{underscore}         
  \usepackage[T1]{fontenc}        
\else
  \usepackage{breakurl}           
\fi

\title{Answer Set Counting and its Applications}
\author{Mohimenul Kabir
\institute{National University of Singapore}
\institute{School of Computing\\
Singapore}
}
\def\titlerunning{Answer Set Counting and its Applications}
\def\authorrunning{Kabir}
\usepackage{amsmath}
\usepackage{amsfonts}
\usepackage{cleveref}
\usepackage{array}
\usepackage{subcaption}
\usepackage{caption}
\usepackage{flexisym}
\usepackage{xfrac}
\usepackage{algorithm}
\usepackage{algpseudocode}
\usepackage{bbm}
\usepackage{xspace}

\usepackage{multirow} 
\usepackage{booktabs}
\usepackage{makecell}

\newcommand{\copyoperation}[1]{\mathsf{Copy}(#1)}
\newcommand{\copyatom}[1]{#1\textprime}
\newcommand{\loopatoms}[1]{\mathsf{LA}(#1)}

\newcommand{\Card}[1]{|#1|}

\newcommand{\true}{\ensuremath{\mathsf{true}}\xspace}
\newcommand{\false}{\ensuremath{\mathsf{false}}\xspace}
\newcommand{\completion}[1]{\mathsf{Comp}(#1)}

\newcommand{\rules}[1]{\mathsf{Rules}(#1)}
\newcommand{\body}[1]{\mathsf{Body}(#1)}
\newcommand{\at}[1]{\mathsf{atoms}(#1)}
\newcommand{\head}[1]{\mathsf{Head}(#1)}
\newcommand{\answer}[1]{\mathsf{AS}(#1)}
\newcommand{\pr}[1]{\mathsf{Pr}[#1]}

\algnewcommand\algorithmicforeach{\textbf{for each}}
\algdef{S}[FOR]{ForEach}[1]{\algorithmicforeach\ #1\ \algorithmicdo}

\begin{document}
\maketitle
\begin{abstract}
  We have focused on Answer Set Programming (ASP), more specifically, answer set counting, exploring both exact and approximate methodologies. We developed an exact ASP counter, sharpASP, which utilizes a compact encoding for propositional formulas, significantly enhancing efficiency compared to existing methods that often struggle with inefficient encodings. Our evaluations indicate that sharpASP outperforms current ASP counters on several benchmarks. In addition, we proposed an approximate ASP counter, named ApproxASP, a hashing-based counter integrating Gauss-Jordan elimination within the ASP solver, clingo. As a practical application, we employed ApproxASP for network reliability estimation, demonstrating superior performance over both traditional reliability estimators and \#SAT-based methods.
\end{abstract}
\section{Introduction}
Answer Set Programming (ASP)~\cite{MT1999} has emerged as a promising paradigm in knowledge representation and automated reasoning owing to its ability to model hard combinatorial problems from diverse domains in a natural way~\cite{brik2015diagnosing}. Building on advances in propositional SAT solving, the past two decades have witnessed the emergence of well-engineered systems for solving the answer set satisfiability problem, i.e., finding models or answer sets for a given answer set program. In recent years, there has been growing interest in problems beyond satisfiability, such as model counting, in the context of ASP. In this work, we focus on the model counting problem in the context of ASP, known as {\em answer set counting} problem. There has been growing interest in answer set counting, motivated by applications in probabilistic reasoning and network reliability~\cite{KM2023, ACMS2015,HJK2022}

\section{Background and Problem Statement}
An \textit{answer set program} $P$ consists of a set of rules, each rule is structured as follows:
\begin{align}
\label{eq:generalrule}
\text{Rule $r$:~~}a_1 \vee \ldots a_k \leftarrow b_1, \ldots, b_m, \textsf{not } c_1, \ldots, \textsf{not } c_n
\end{align}
where, $a_1, \ldots, a_k, b_1, \ldots, b_m, c_1, \ldots, c_n$ are propositional variables or atoms, and $k,m,n$ are non-negative integers. 
The notations $\rules{P}$ and $\at{P}$ denote the rules and atoms within the program $P$. 
In rule $r$, the operator ``\textsf{not}'' denotes \textit{default negation}~\cite{clark1978}. For each 
rule $r$ (\cref{eq:generalrule}), we adopt the following notations: the atom set $\{a_1, \ldots, a_k\}$ constitutes the {\em head} of $r$, denoted by $\head{r}$, the set $\{b_1, \ldots, b_m\}$ is referred to as the {\em positive body atoms} of $r$, denoted by $\body{r}^+$, and the set $\{c_1, \ldots, c_n\}$ is referred to as the \textit{negative body atoms} of $r$, denoted by $\body{r}^-$.
A rule $r$ is called a {\em constraint} when $\head{r}$ contains no atom.
A program $P$ is called a {\em disjunctive logic program} if there is a rule $r \in \rules{P}$ such that $\Card{\head{r}} \geq 2$~\cite{BD1994}.

In ASP, an interpretation $M$ over $\at{P}$ specifies which atoms are assigned \true; that is, an atom $a$ is \true under $M$ if and only if $a \in M$ (or \false when $a \not\in M$ resp.). 
An interpretation $M$ satisfies a rule $r$, denoted by $M \models r$, if and only if $(\head{r} \cup \body{r}^{-}) \cap M \neq \emptyset$ or $\body{r}^{+} \setminus M \neq \emptyset$. An interpretation $M$ is a {\em model} of $P$, denoted by $M \models P$, when $\forall_{r \in \rules{P}} M \models r$. 
The \textit{Gelfond-Lifschitz (GL) reduct} of a program $P$, with respect to an interpretation $M$, is defined as $P^M = \{\head{r} \leftarrow \body{r}^+| r \in \rules{P}, \body{r}^- \cap M = \emptyset\}$~\cite{GL1991}.
An interpretation $M$ is an {\em answer set} of $P$ if $M \models P$ and no $M\textprime \subset M$ exists such that $M\textprime \models P^M$.
We denote the answer sets of program $P$ using the notation $\answer{P}$.

\paragraph{Exact Answer Set Counting~\cite{KCM2024}}
Given an ASP program $P$, the exact answer set counting seeks to count the number of answer sets of $P$; more formally, the problem seeks to find $\Card{\answer{P}}$.

\paragraph{Approximate Answer Set Counting~\cite{KESHFM2022}}
Given an ASP program $P$, tolerance parameter $\epsilon$, and confidence parameter $\delta$, the approximate answer set counting seeks to estimate the number of answer sets of $P$ with a probabilistic guarantee; more formally, the approximate answer set counters returns a count $c$ such that 
$\pr{\sfrac{\Card{\answer{P}}}{1 + \epsilon} \leq c \leq (1 + \epsilon) \times \Card{\answer{P}}} \geq 1 - \delta$. Our approximate answer set counter invokes a polynomial number of calls to an ASP solver. 

\paragraph{Clark's completion} \cite{clark1978} or \emph{program
completion} is a technique to translate a normal program $P$ into a propositional formula $\completion{P}$ that is related but not semantically equivalent. Specifically, for each atom $a$ in $\at{P}$, we perform the following steps:
\begin{enumerate}
    \item Let $r_1, \ldots, r_k \in \rules{P}$ such
      that $\head{r_1} =\ldots= \head{r_k} = a$, then we add the
      propositional formula $(a \leftrightarrow (\body{r_1} \vee
      \ldots \vee \body{r_k}))$ to $\completion{P}$.
    \item Otherwise, we add the literal $\neg{a}$ to $\completion{P}$.
\end{enumerate}
Finally, $\completion{P}$ is derived by logically conjoining all the previously added constraints. Literature indicates that while every answer set of $P$ satisfies $\completion{P}$, the converse is not true~\cite{LZ2004}.
\section{Related Works}
\label{section:relatedwork}
The decision version of normal logic programs is NP-complete; therefore, the ASP counting for normal logic programs is \#P-complete~\cite{valiant1979} via a polynomial reduction~\cite{JN2011}.  Given the \#P-completeness, a prominent line of work focused on ASP counting relies on translations from the ASP program to a CNF formula~\cite{LZ2004,Janhunen2004,Janhunen2006,JN2011}. Such translations often result in a large number of CNF clauses and thereby limit practical scalability for {\em non-tight} ASP programs. 
Eiter et al.~\cite{EHK2024} introduced T$_{\mathsf{P}}$-\textit{unfolding} to break cycles and produce a tight program. They proposed an ASP counter called aspmc, that performs a treewidth-aware Clark completion from a cycle-free program to a CNF formula. Jakl, Pichler, and Woltran~\cite{JPW09} extended the tree decomposition based approach for \#SAT due to Samer and Szeider~\cite{SS2010} to ASP and proposed a {\em fixed-parameter tractable} (FPT) algorithm for ASP counting. 
Fichte et al.~\cite{FHMW2017,FH2019} revisited the FPT algorithm due to Jakl et al.~\cite{JPW09}  and developed an exact model counter, called DynASP, that performs well on instances with {\em low treewidth}. 
Aziz et al.~\cite{ACMS2015} extended a propositional model counter to an answer set counter by integrating unfounded set detection. 
ASP solvers~\cite{GKS2012} can count answer set via enumeration, which is suitable for a sufficiently small number of answer sets.
Kabir et al.~\cite{KESHFM2022} focused on lifting hashing-based techniques to ASP counting, resulting in an approximate counter, called ApproxASP, with $(\varepsilon,\delta)$-guarantees. 
Kabir et al.~\cite{KCM2024} introduced an ASP counter that utilizes a sophisticated Boolean formula, termed the copy formula, which features a compact encoding.

\section{Current Progress and Future Goals}
We have already engineered two ASP counters: SharpASP~\cite{KCM2024} and ApproxASP~\cite{KESHFM2022}. 
SharpASP\footnote{\url{https://github.com/meelgroup/SharpASP}} is an exact answer set counter and ApproxASP\footnote{\url{https://github.com/meelgroup/ApproxASP2}} is an approximate answer set counter. 

The principal contribution of SharpASP is to design a scalable answer set counter, without a substantial increase in the size of the transformed propositional formula, particularly when addressing circular dependencies. The key idea behind a substantial reduction in the size of the transformed formula is an alternative yet correlated perspective of defining answer sets. This alternative definition formulates the answer set counting problem on a pair of Boolean formulas $(F, G)$, where the formula $F$ over-approximates the search space of answer sets, while the formula $G$ exploits {\em justifications} to identify answer sets correctly. We set $F = \completion{P}$ since every answer set satisfies Clark completion. Note that $\completion{P}$ overapproximates answers sets of $P$. We propose another formula, named {\em copy formula}, denoted as $\copyoperation{P}$, which comprises a set of (implicitly conjoined) implications defined as follows:
\begin{enumerate}
\item \label{l1:type1} (type 1) for every $v \in \loopatoms{P}$, the implication $\copyatom{v} \rightarrow v$ is in $\copyoperation{P}$.
\item \label{l1:type2} (type 2) for every rule $x \leftarrow a_1, \ldots a_k, b_1, \ldots b_m, \sim c_1, \ldots \sim c_n$ in $P$, where
  $x \in \loopatoms{P}$,
  $\{a_1, \ldots a_k\} \subseteq \loopatoms{P}$ and
  $\{b_1, \ldots b_m\} \cap \loopatoms{P} = \emptyset$,
  the implication $\copyatom{a_1} \wedge \ldots \copyatom{a_k} \wedge b_1 \wedge 
  \ldots b_m \wedge \neg{c_1} \wedge \ldots \neg{c_n} \rightarrow \copyatom{x}$ is in $\copyoperation{P}$.
\item No other implication is in $\copyoperation{P}$.
\end{enumerate}

For each satisfying assignment $M \models \completion{P}$, we have the following observations:
\begin{itemize}
    \item if $M \in \answer{P}$, then $\copyoperation{P}_{|M} = \emptyset$
    \item if $M \not\in \answer{P}$, then $\copyoperation{P}_{|M} \neq \emptyset$
\end{itemize}
where $\copyoperation{P}_{|M}$ denotes the {\em unit propagation} of $M$ on $\completion{P}$. We integrate these observations into propositional model counters to engineer an answer set counter.


Within ApproxASP, we present a scalable approach to approximate the number of answer sets.
Inspired by approximate model counter ApproxMC~\cite{CMV2013}, our approach is based on systematically adding parity (XOR) constraints to
ASP programs to divide the search space uniformly and independently. We prove that adding random XOR constraints partitions the 
answer sets of an ASP program. 
When a randomly chosen partition is {\em quite small}, we can approximate the number of answer sets by simple multiplication. 
The XOR semantic in answer set programs was initiated by Everardo et al.~\cite{EJKS2019}.
In practice, we use a {\em Gaussian
elimination}-based approach by lifting ideas from SAT to ASP and
integrating them into a state-of-the-art ASP solver.

Our objective is to develop more efficient answer set counters by integrating specialized ASP counting techniques and advanced preprocessing methods. Furthermore, we are dedicated to enhancing the capabilities of SharpASP, currently limited to handling normal programs, to also support disjunctive answer set programs. In addition, we are eager to explore broader applications of ASP counting to demonstrate its versatility and potential in solving complex problems. We are also eager to extend the counting technique in other theories~\cite{KM2024}.
\section{Some Results}
We implemented prototypes of both SharpASP, on top of the existing propositional model counter SharpSAT-TD (denoted as SharpASP (STD) and ApproxASP, on top of ASP solver Clingo. Finally, we empirically evaluate their performance against existing counting benchmarks used in answer set counting literature~\cite{FHMW2017,EHK2024,ACMS2015}.

\paragraph{SharpASP}
Our extensive empirical analysis of $1470$ benchmarks demonstrates significant performance gain over current state-of-the-art exact answer set counters. The result demonstrated is presented in Table~$1$ and the rightmost column presents the result of SharpASP. Specifically, by using SharpASP, we were able to solve $1023$ benchmarks with a PAR$2$ score of $3373$, whereas using prior state-of-the-art, we could only solve $869$ benchmarks with a PAR$2$ score of $4285$. A detailed experimental analysis revealed that the strength of SharpASP is that it spends less time in binary constraint propagation while making more decisions compared to off-the-shelf propositional model counters.

\begin{table}[t]
      \centering
      \begin{tabular}{m{4em} m{4em} m{4em} m{4em} m{4em} m{4em}}
      \toprule
      & clingo & ASProb & aspmc+STD & lp2sat+STD & SharpASP (STD)\\
     \midrule 
      \#Solved  & 869 & 188 & 840 & 776 & \textbf{1023}\\
      \midrule
      PAR$2$ & 4285 & 8722 & 4572 & 5084 & \textbf{3373}\\
      \bottomrule
      \end{tabular}
      \caption{The performance comparison of SharpASP vis-a-vis other ASP counters in terms of the number of solved instances and PAR$2$ scores.
      }
\end{table}

\paragraph{ApproxASP}
Table~$2$ presents the result of ApproxASP with state-of-the-art answer set counters.
ApproxASP performs well in disjunctive logic programs.
ApproxASP solved $185$ instances among $200$ instances, while the
best ASP solver clingo solved a total of $177$ instances.  In
addition, on normal logic programs, ApproxASP performs on par with
state-of-the-art approximate model counter ApproxMC.

\begin{table}[t]
\centering
\begin{tabular}{m{1em} m{4em} m{4em}  m{4em}  m{4em}  m{4em} m{4em} } 
\toprule
&& Clingo & DynASP & Ganak & ApproxMC & ApproxASP\\
\midrule
\multirow{3}{*}{\rotatebox{90}{Normal}} 
& \#Instances & \multicolumn{5}{|c|}{1500}\\\cmidrule{2-7}
& \#Solved & 738 & 47 & 973 & \textbf{1325} & 1323\\
\cmidrule{2-7}
&PAR-$2$ & 5172 & 9705 & 3606 & \textbf{1200} & 1218\\
\midrule
\multirow{3}{*}{\rotatebox{90}{Disjunc.}} 
& \#Instances & \multicolumn{5}{|c|}{200}\\
\cmidrule{2-7}
& \#Solved & 177 & 0 & 0 & 0 & \textbf{185}\\
\cmidrule{2-7}
& PAR$2$ & 1372 & 10000 & 10000 & 10000 & \textbf{795}\\
\bottomrule
\end{tabular}
\caption{%
  The runtime performance comparison of Clingo, DynASP, Ganak, ApproxMC, and ApproxASP on all considered instances.
}
\end{table}


\section{Open issues and expected achievements}
Model counting, an intractable problem, is classified as \#P for normal programs and \#co-NP
\cite{FHMW2017} for disjunctive logic programs, presenting significant challenges in developing scalable answer set counters. Our observations indicate that while our engineered counters effectively scale for certain problem types, they underperform for others. The diverse applications of model counting in real-world scenarios further complicate the creation of application-specific ASP counters. Moreover, we have identified instances where existing systems outperform our SharpASP counter. Integrating strengths from these existing counters into SharpASP to enhance its scalability remains a formidable challenge.

\bibliographystyle{eptcs}
\bibliography{example}
\end{document}